\documentclass[11pt]{article}

\usepackage[final]{acl}

\usepackage{times}
\usepackage{latexsym}

\usepackage[T1]{fontenc}

\usepackage[utf8]{inputenc}

\usepackage{microtype}

\usepackage{inconsolata}

\usepackage{graphicx}
\usepackage{booktabs}
\usepackage{amsmath}
\usepackage{amssymb}
\usepackage{pifont}
\usepackage[table]{xcolor}
\usepackage{multirow}
\usepackage{tabularx}
\usepackage{url}
\usepackage{float}

\title{CodecCap: High-Fidelity Codec-Inspired Residual Modeling for \\ Dense Video Captioning}

\author{
  Zihan Lin$^{1}$\thanks{Equal contribution.}, Songhe Deng$^{1}$\footnotemark[1], Shuwei He$^{1,2}$\footnotemark[1], Danxiang Zhu$^1$, Dan Zhang$^1$, \\ \textbf{Yishu Lei}$^1$, \textbf{Xianlong Luo}$^1$, 
  \textbf{Shikun Feng}$^1$\thanks{Corresponding author.}, \textbf{Rui Liu}$^{2}$ \\
  $^1$~ERNIE Team, Baidu \quad
  $^2$~College of Artificial Intelligence, Inner Mongolia University \\
  \texttt{\{linzihan04, dengsonghe, heshuwei, zhudanxiang, zhangdan20, leiyishu,}  \\
  \texttt{luoxianlong, fengshikun01\}@baidu.com, imucslr@imu.edu.cn}
}

\begin{document}
\maketitle

\begin{abstract} 
Existing video captioning methods struggle to balance visual fidelity and redundancy: holistic captions are compact but lose fine-grained evidence, whereas segment-wise captions improve coverage but  introduce heavy redundancy.
We propose \textbf{CodecCap}, a codec-inspired framework for high-fidelity dense video captioning.
Analogous to video codecs, CodecCap represents videos using keyframe and residual captions. Keyframe captions exhaustively encode stable visual context, while residual captions capture temporally only localized actions, motions and changes. 
This effectively preserves fine-grained visual evidence while reducing redundant descriptions.
To quantify the fidelity of captions, we introduce \textbf{VidCapQA}, a caption-then-QA benchmark with 1,000 questions across 14 capability dimensions. Results on VidCapQA show that captions directly generated by strong VLMs still miss many visual details, highlighting caption representation as a critical bottleneck.
Experiments show that CodecCap significantly surpass direct captioning with the same underlying VLMs, suggesting keyframe–residual captioning an way for high-fidelity video-language supervision.
We further use CodecCap to construct CodecVDC-100K, a large-scale dense captioning dataset with anchor, residual, scene-level, and video-level supervision.
\end{abstract}
\section{Introduction}

Recent progress in visual-language modeling has increasingly expanded from static images to the video domain, enabling strong performance in video captioning~\citep{zhang2025tarsier2}, video understanding~\citep{chen2024sharegpt4video,yang2024vript}, and question answering~\citep{mvbench,videomme}. 
As a fundamental video-language task, video captioning converts visual dynamics into language and provides an important source of supervision for video-language learning. 
However, generic video-level captions often provide only coarse summaries, overlooking temporally localized events and fine-grained visual changes. 
Video Dense Captioning~(VDC) addresses this limitation by requiring captions that are temporally grounded and visually faithful, capturing local actions, state changes, event ordering, and object interactions while avoiding redundant descriptions of persistent scenes or actors. 
Constructing such captions at scale remains challenging, as annotation pipelines must preserve fine-grained visual evidence while suppressing redundant or weakly grounded descriptions.

\begin{figure}[t]
\centering
\includegraphics[width=\linewidth, trim=0.4cm 0.4cm 0.4cm 0.4cm, clip]
{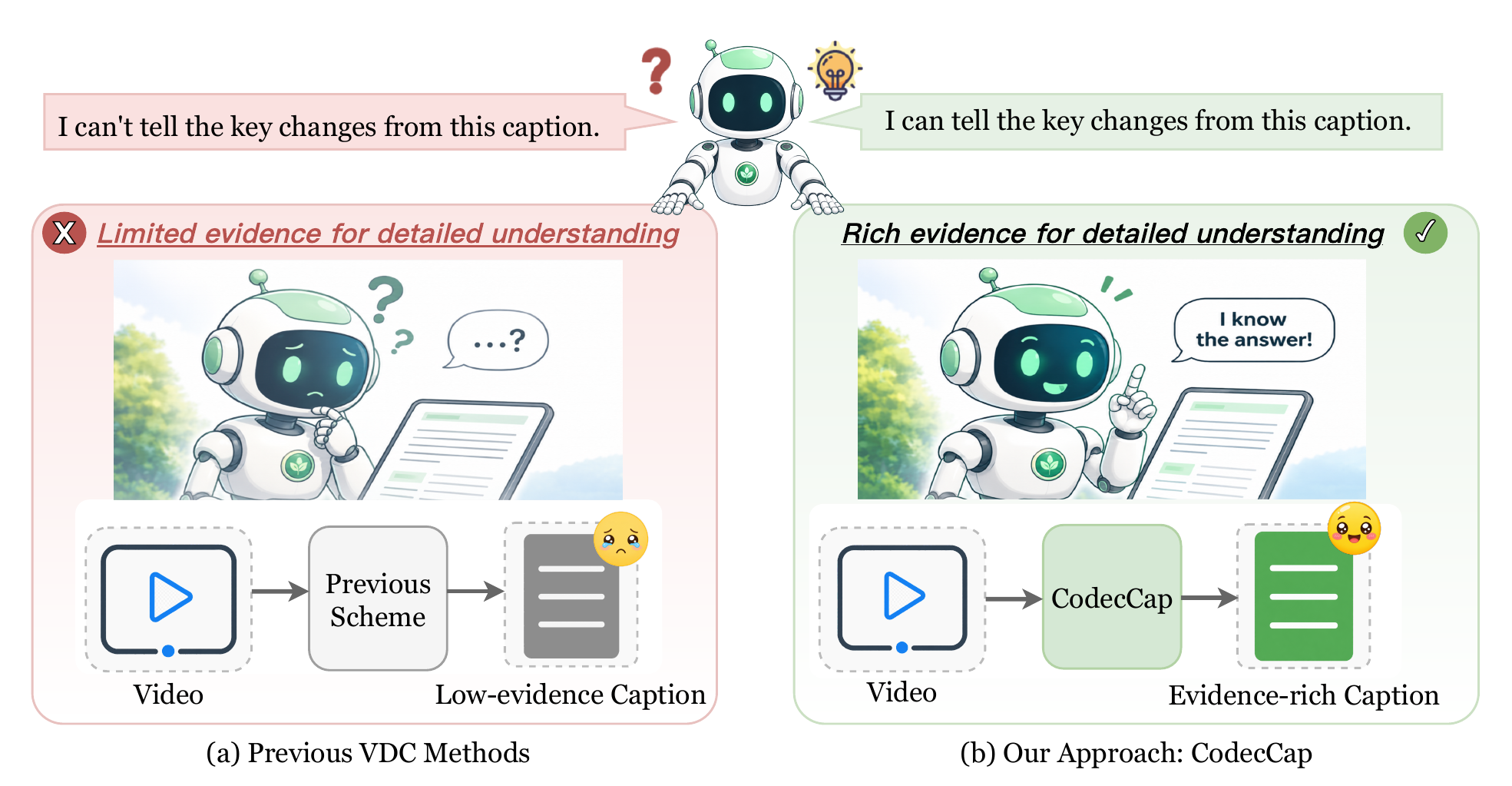}
\caption{
Previous VDC methods often omit critical changes or repeatedly describe stable content, whereas CodecCap anchors persistent content once and captures changes with compact residual descriptions.
}
\label{fig:overview}
\end{figure}

Existing video captioning pipelines typically represent video content either holistically or segment-wise.
Holistic captions are compact but often miss temporally localized events and fine-grained visual changes~\citep{zhang2025tarsier2,vdc_agent}.
Segment-wise or dense descriptions improve temporal coverage~\citep{krishna2017dense,mun2019streamlined,ta_prompting}, but tend to repeat persistent scenes, objects, and motions across neighboring segments. 
This leads to a clear dilemma: captions are often either compact but evidence-poor, or detailed but redundant. 
Moreover, this dilemma is poorly captured by conventional captioning metrics, which emphasize semantic similarity over whether fine-grained visual events can be recovered from the caption. 
These limitations motivate us to first diagnose the recoverability of visual evidence from captions, and then design a more structured caption representation that preserves temporal changes without repeatedly describing stable content.

We first propose \textbf{VidCapQA}, a question-answering-based evaluation protocol that probes whether generated captions retain the visual evidence needed to answer questions of the original video.
Instead of directly evaluating VLMa video-language model on videos, VidCapQA providing only the generated caption to an text-only LLM, which then answers multiple-choice questions about the original video.
The questions in VidCapQA are sampled from eight high-quality video understanding benchmarks~\citep{mvbench,motionbench,tempcompass,tomato,etbench,longvideobench,lvbench,videomme}. We further re-label them into 14 capability dimensions and construct a stratified 1{,}000-question evaluation set to provide balanced diagnostic coverage. 
Under this protocol, captions generated by strong models such as Gemini 3.1 Pro~\cite{gemini31pro} achieve only 51.8\% accuracy, suggesting that current caption representations still lose substantial recoverable visual evidence.

To address this limitation, we propose \textbf{CodecCap}, a codec-inspired dense captioning framework. 
Modern video codecs~\cite{av1,h264} efficiently represent visual streams by exploiting temporal redundancy: I-frames encode complete visual content, whereas P-frames store only the residual changes with respect to previous frames.
Inspired by this principle, CodecCap rethinks video captioning from a semantic coding perspective. 
Instead of generating repetitive descriptions for each frame or compressing an entire video into a single caption, CodecCap assigns comprehensive and stable descriptions to keyframe anchors, serving as semantic I-frames, and uses incremental residual captions to capture newly emerging actions, motions, state changes, and interactions, serving as semantic P-frames.
Concretely, CodecCap segments videos into scene-aligned units and pairs each unit with a keyframe anchor and per-second residual captions. This decomposition turns video captioning from holistic summarization into structured semantic representation: persistent scene context is recorded once, while temporally localized visual changes are captured explicitly.

CodecCap is model-agnostic and can be applied to different VLMs. 
Using the same underlying model as the direct-captioning baseline, CodecCap improves performance on VidCapQA, indicating that the gains stem from the proposed keyframe–residual representation rather than increased model capacity alone.
Its structured design also enables scalable dense caption synthesis. Based on CodecCap, we construct \textbf{CodecVDC-100K}, a large-scale video dense captioning dataset containing 100K+ videos and 7{,}400+ hours of video content. Extensive experiments show that CodecCap-generated supervision improves caption recoverability on VidCapQA over direct-captioning baselines with the same underlying VLMs. These results demonstrate that codec-style semantic residual modeling provides an effective and scalable representation for preserving temporally grounded visual evidence in video captions.

In summary, our main contributions are:
\begin{itemize}
    \item We introduce \textbf{CodecCap}, a codec-inspired captioning framework that represents videos with semantic keyframes and residual captions, preserving fine-grained temporal changes while reducing redundancy.
    \item We propose \textbf{VidCapQA}, a QA-based diagnostic benchmark that evaluates whether video captions preserve recoverable visual evidence, with 1{,}000 questions in 14 capability dimensions.
    \item We construct and will release \textbf{CodecVDC-100K}, a 100K+-video, 7{,}400+-hour dense captioning dataset with four-level supervision, and show that CodecCap consistently outperforms direct captioning with the same underlying VLMs.
\end{itemize}
\section{Related Work}

\subsection{Video Dense Captioning}

Video dense captioning has evolved from holistic captioning to segment-wise, hierarchical, and structured generation paradigms.
Holistic methods~\citep{zhang2025tarsier2, vdc_agent, chai2024auroracap} generate a single video-level description but often omit local evidence as the video span grows.
Segment-wise strategies~\citep{krishna2017dense, mun2019streamlined, wang2022end2end, ta_prompting} preserve local detail by captioning clips independently, but repeatedly describe stable content.
Hierarchical and structured approaches~\citep{han2024arc, wang2025hivid, li2024wolf, mr_video, yang2024vript, chen2024sharegpt4video} organize or factorize descriptions across granularities, yet often duplicate content or collapse residual structure at inference time.
Overall, existing methods remain constrained by a trade-off between evidence preservation and redundancy control.
In contrast, CodecCap addresses this limitation by retaining a four-level caption hierarchy (keyframe / residual / per-scene / whole-video) at deployment time.

\subsection{Efficient Video-Text Representation}

Modern video codecs, including H.264/AVC~\citep{h264}, H.265/HEVC~\citep{hevc}, and AV1~\citep{av1}, exploit temporal redundancy with intra-coded frames~(I-frames) and predictive or bidirectional frames~(P/B-frames).
CodecCap transfers this principle to VDC, using keyframe descriptions for stable content and textual residuals for changes.
Prior work on long-video efficiency reduces input-side cost through hierarchical compression~\citep{wang2025hivid}, streaming architectures~\citep{streaming_vit, livecc2025}, retrieval-augmented routing~\citep{salova}, token-efficient encoding~\citep{chen2025storm, videoloom}, or learned video compression~\citep{tempo2026}.
Change-focused description has also been studied for remote sensing images~\citep{changeimti}, video pairs~\citep{wu2026vidic}, and differential annotation~\citep{chen2024sharegpt4video}.
Unlike these input-compression or pairwise-difference settings, CodecCap maintains a four-level caption hierarchy (keyframe / residual / per-scene / whole-video) for a single video at inference time.

\subsection{Caption Quality Evaluation}

Evaluating whether dense captions preserve evidence for downstream reasoning remains challenging.
N-gram metrics, such as BLEU~\citep{bleu}, CIDEr~\citep{cider}, and METEOR~\citep{meteor}, rely on lexical overlap and correlate poorly with factuality judgments~\citep{liu2023factvc}.
Factuality and hallucination metrics~\citep{chaturvedi2024ovfact, jha2025argus, choong2026vidhal, nakada2025hlvc, lee2025noah} verify unfaithful claims but often require references or perturbations.
Question-answering or multiple-choice evaluation~\citep{zhang2025ifvidcap, xiao2024hmgie} assesses captions through questions but usually depends on per-video question synthesis or manual authoring.
Information-preservation metrics~\citep{visil2026, kong2025tuna} provide scalar scores without capability-specific diagnosis, while video-understanding benchmarks~\citep{mvbench, motionbench, tempcompass, tomato, etbench, longvideobench, lvbench, videomme} evaluate VLMs rather than captions.
VidCapQA addresses this gap with a caption-then-predict protocol, where a text-only LLM answers community-validated questions conditioned only on the generated caption.
\section{CodecCap: Methodology}
\label{sec:method}

\begin{figure*}[t]
\centering
\includegraphics[width=\linewidth, trim=0.3cm 0.3cm 0.3cm 0.3cm, clip]
{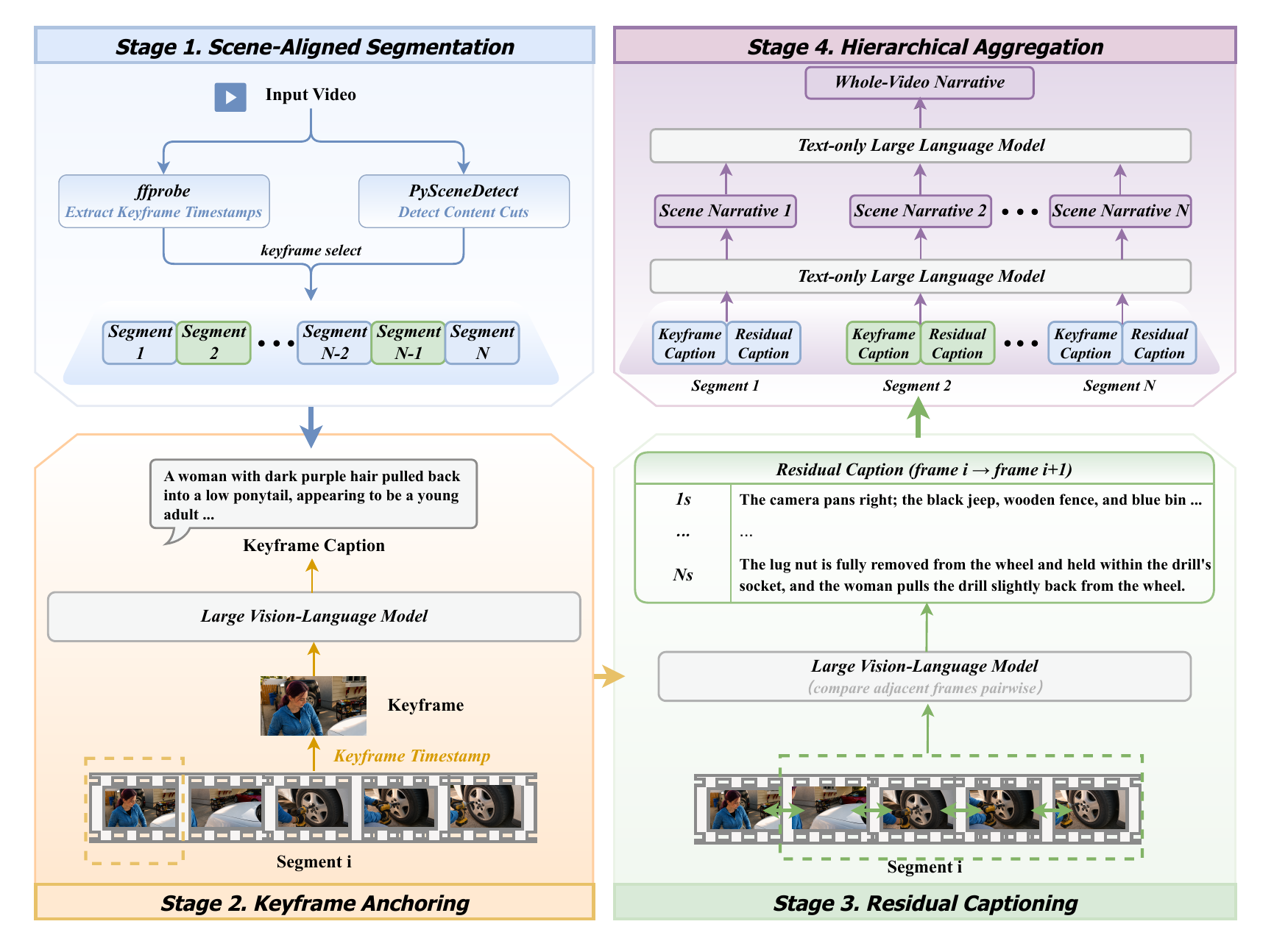}
\caption{
Overview of CodecCap. The pipeline segments a video into scene-aligned segments, encodes each segment with a high-fidelity semantic anchor and temporal residual sequences, then aggregates the resulting evidence into scene-level and whole-video narratives.
}
\label{fig:codeccap_pipeline}
\end{figure*}

\subsection{Task Formulation and Overview}
\label{sec:task_def}

Given a video $V=\{f_t\}_{t=1}^{T}$, dense video captioning produces a structured textual representation $\mathcal{C}=\mathcal{F}(V)$ that preserves visual evidence without redundant description.
Inspired by modern video codecs~\citep{h264, hevc}, CodecCap factorizes each scene-aligned segment into a full-state \emph{anchor} caption and temporal \emph{residual} captions: the anchor records the stable visual state (scene, entities, attributes, spatial layout, etc.), while residuals encode only subsequent changes (motion, state transitions, interactions, etc.).

Formally, $\mathcal{C}$ comprises four granularities: keyframe captions $\{c^{\mathrm{anc}}_k\}_{k=1}^{K}$, per-second residual captions $\{\Delta c_{k,i}\}_{i=1}^{N_k}$, scene-level narratives $\{c^{\mathrm{scene}}_k\}_{k=1}^{K}$, and a whole-video narrative $c^{\mathrm{vid}}$.
As shown in Figure~\ref{fig:codeccap_pipeline}, CodecCap constructs this representation through scene-aligned segmentation, anchor captioning, contextual residual captioning, and hierarchical evidence aggregation.
A vision-language model (VLM) generates visual captions with structured prompts; a text-only LLM validates and synthesizes the evidence.

\subsection{Scene-Aligned Segmentation}
\label{sec:scene_seg}

CodecCap first partitions videos into coherent segments by combining codec metadata with content-based shot detection.
In design, video codecs inserts an I-frame when scene changes. But it also imposes a maximum GOP (Group of Pictures) length to support random access and bound error propagation, causing I-frames to appear periodically even within continuous scenes.
These \emph{codec-refresh} I-frames carry no semantic boundary information, whereas I-frames in edited or variable-bitrate videos often coincide with actual scene cuts.

CodecCap distinguishes these regimes with an adaptive GOP-aware strategy.
It extracts I-frame timestamps using \texttt{ffprobe}\footnote{\url{https://ffmpeg.org}} and detects content cuts using PySceneDetect~\citep{pyscenedetect}.
The coefficient of variation (CV) of inter-I-frame gaps is used to measures I-frame regularity.
When $\text{CV} \geq \tau_{\text{gop}}$, CodecCap uses \emph{I-frame-primary} mode: I-frames are candidate boundaries retained only if a PySceneDetect cut lies within a temporal proximity window.
When $\text{CV} < \tau_{\text{gop}}$, indicating fixed-GOP encoding (e.g., live streams, HLS recordings), CodecCap uses \emph{content-primary} mode and takes PySceneDetect cuts directly as boundaries.
To prevent very long segments from exceeding the temporal context used by downstream captioning, CodecCap further enforces a maximum segment duration.
The resulting segments are both content-coherent and bounded in length, providing stable units for anchor-residual captioning and later aggregation.

\subsection{Keyframe Anchoring}
\label{sec:keyframe_anchor}

For each segment, the first sampled frame serves as the anchor, ensuring that all residuals are forward differences from a known reference.
Because segments begin at detected scene transitions, this frame captures the new scene's major stable content.
The VLM produces an exhaustive anchor caption that covers the environment, entities, visual attributes, object states, and spatial layout.
This anchor serves as visual anchoring to avoid repeatedly describing low-entropy static content in subsequent frames.
During aggregation, static properties from the anchor are preserved by default and are updated only when residual evidence explicitly indicates a change.

\subsection{Residual Captioning}
\label{sec:residual_cap}

Residual captioning focuses on visual changes between adjacent sampled frames while avoiding repeated descriptions of unchanged content.
For each segment, CodecCap samples frames at a fixed rate and processes them in temporal windows.

\textbf{Delta-only enforcement.}
Within each window, The VLM output each residual as a \texttt{\{frame\_pair, delta\_caption\}} JSON object, forcing atomic change descriptions rather than scene re-summaries.
Each residual caption is constrained to describe only visible changes relative to the previous frame: object displacement, motions, interactions, occlusion changes, etc.

\textbf{Multi-frame context.}
Pairwise frame comparison is sensitive to compression artifacts, occlusion, and camera jitter.
CodecCap therefore introduces a sliding-window design by giving the VLM a window of consecutive frames, allowing it to distinguish persistent changes from transient noise. Adjacent windows share boundary frames for continuity.

\textbf{Spatial normalization.}
To keep consistency between each description, CodecCap normalizes the spatial language as follows:
Positions use a coarse 9-zone grid (e.g., upper-left, center, lower-right) and, when finer precision is needed, normalized coordinates $(x\%, y\%)$.
Directional terms follow screen coordinates rather than subject-centric ones, enabling consistent trajectory reconstruction during aggregation.

\begin{figure*}[t]
   \centering
\includegraphics[width=1.0\linewidth]
{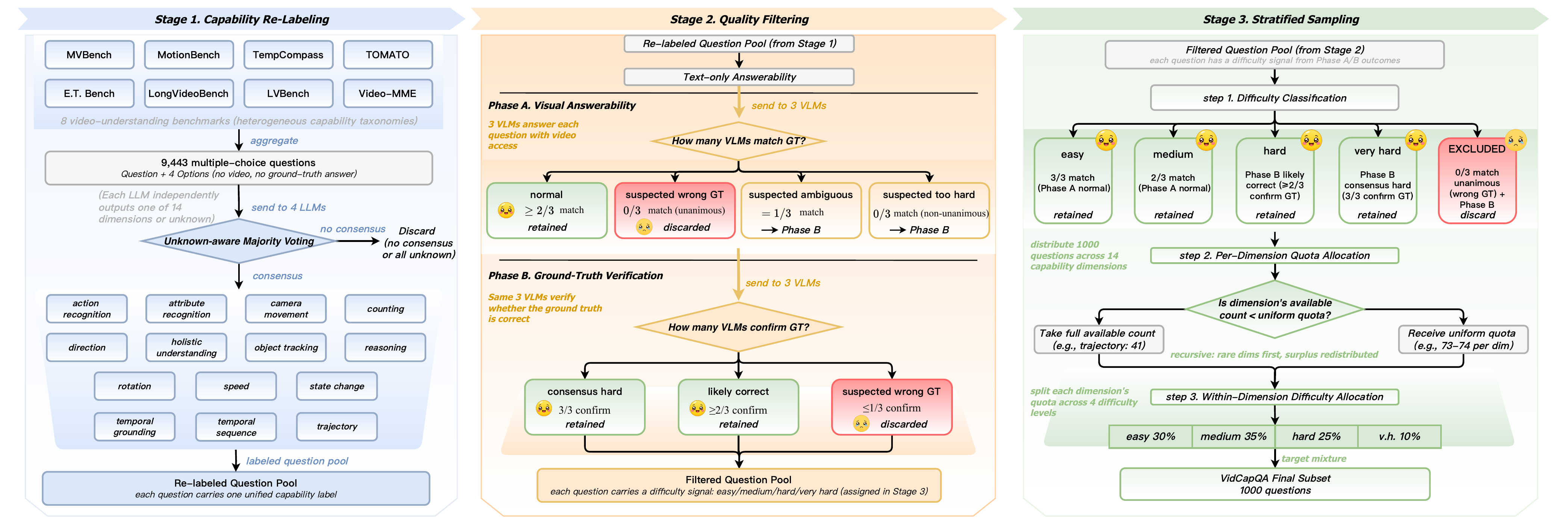}
\caption{
Construction pipeline of VidCapQA. Questions from eight video-understanding benchmarks are re-labeled into 14 capability dimensions, filtered through text-only and visual quality checks, and stratified-sampled into a balanced 1{,}000-question subset.
}
\label{fig:vidcapqa_pipeline} 
\end{figure*}

\subsection{Hierarchical Aggregation}
\label{sec:aggregation}
The residual stream provides dense temporal evidence, but is intentionally fragmentary.
CodecCap further aggregates anchor and residual captions into coherent scene-level and video-level narratives using a text-only LLM.

Before synthesis, the LLM applies structured acceptance rules to filter hallucinations and artifacts:
\begin{itemize}
  \item \textbf{Continuous changes} require support from at least two consecutive residuals.
  \item \textbf{Discrete events} may be accepted from one residual if the before/after context is consistent.
  \item \textbf{Static attributes} of the anchor persist unless a residual explicitly reports a change (attribute locking).
  \item \textbf{Contradictions} are resolved by support count; unresolved conflicts are omitted.
\end{itemize}

After validation, the LLM composes the evidence into a fluent scene-level paragraph and avoids unsupported inferences.
Whole-video synthesis concatenates scene narratives chronologically with the same rules applied across segment boundaries.
The final representation exposes four levels of detail---anchor captions, per-second residual captions, scene narratives, and a whole-video narrative---allowing downstream models or evaluators to select the granularity required by each task.

\section{VidCapQA: Benchmark}
\label{sec:vidcapqa}

\subsection{Benchmark Overview}

VidCapQA evaluates whether generated dense captions preserve the evidence required for capability-specific video understanding.
Unlike lexical-overlap metrics~\citep{bleu,cider,meteor}, factuality metrics~\citep{chaturvedi2024ovfact,jha2025argus,lee2025noah}, or end-to-end VLM evaluation~\citep{mvbench,videomme}, VidCapQA isolates caption quality through a caption-then-predict protocol~\citep{visil2026,zhang2025ifvidcap}: a captioner first describes a video, and a text-only LLM answers the question using only the generated caption.
As shown in Figure~\ref{fig:vidcapqa_pipeline}, VidCapQA is constructed in three stages: capability re-labeling, quality filtering, and stratified sampling.

\subsection{Capability Re-Labeling}
\label{sec:vidcapqa_pool}

VidCapQA aggregates 9{,}443 multiple-choice questions from eight video-understanding benchmarks~\citep{mvbench,motionbench,tempcompass,tomato,etbench,longvideobench,lvbench,videomme}; per-source statistics are provided in Appendix~\ref{app:vidcapqa_sources}.
Because these benchmarks use heterogeneous taxonomies, we re-label each question into one of 14 capability dimensions: action recognition, attribute recognition, camera movement, counting, direction, holistic understanding, object tracking, reasoning, rotation, speed, state change, temporal grounding, temporal sequence, and trajectory.
Four LLMs---Claude Sonnet 4.6, o4-mini, Doubao-Seed 2.0-Pro, and Gemini-3 Flash-Preview---independently assign a label from the question and answer options, without access to the video or ground-truth answer.
We accept a label only when it is supported by at least two LLMs and preferred over an unknown fallback; otherwise, the question is discarded.
We refer to this consensus rule as \emph{unknown-aware majority voting}.

\subsection{Quality Filtering}
\label{sec:vidcapqa_filter}

We apply a sequence of filters to remove text-leakable, mislabeled, ambiguous, or unanswerable questions.
We first run a \emph{text-only answerability filter}~\citep{videomme} to ensure that retained questions genuinely require visual evidence: two strong text-only language models---GPT-OSS-120B\footnote{\url{https://huggingface.co/openai/gpt-oss-120b}} and Gemini-3-Flash\footnote{\url{https://ai.google.dev/gemini-api}}---independently attempt each question conditioned only on the question text and four options, without access to the video; questions answered correctly by either model are discarded as text-leakable.
We then apply a two-phase visual filter using video-aware VLMs.

In Phase~A, three VLMs---Doubao-Seed 2.0-Pro, Gemini-3 Pro-Preview, and Qwen3.6-Plus---answer each question with video access.
We classify each question by the number of VLMs whose answer matches the ground truth: questions with $\geq\!2/3$ matches are retained as \emph{normal}; questions with $0/3$ matches under unanimous responses are discarded as \emph{suspected wrong ground truth}; the remaining cases ($1/3$ match, or $0/3$ under non-unanimous responses) are passed to Phase~B.

In Phase~B, the same three VLMs are re-prompted to verify whether the original ground-truth answer is correct given the video.
We retain a question as \emph{consensus hard} when all $3/3$ VLMs confirm the ground truth, as \emph{likely correct} when $\geq\!2/3$ confirm, and discard otherwise.
This procedure preserves difficult but valid questions while removing label noise unresolved by direct answering.

\subsection{Stratified Sampling}
\label{sec:vidcapqa_sample}

From the filtered pool, we sample a stratified 1{,}000-question subset balanced across 14 capability dimensions and four difficulty levels.
Difficulty is assigned from the Phase-A and Phase-B outcomes: questions answered correctly by $3/3$ VLMs in Phase~A are \emph{easy}; $2/3$ correct are \emph{medium}; Phase-B \emph{likely correct} questions are \emph{hard}; Phase-B \emph{consensus hard} questions are \emph{very hard}.
We first allocate the overall budget across capability dimensions, giving rare dimensions their full available counts and redistributing surplus to higher-resource dimensions.
Within each dimension, we allocate questions according to the target mixture of 30\% easy, 35\% medium, 25\% hard, and 10\% very hard.
The final subset contains 73--74 questions for most dimensions and 41 questions for trajectory, the rarest dimension.

\subsection{Evaluation Protocol}
\label{sec:vidcapqa_eval}

For each video $v_i$, a captioner generates a caption $C_i$.
A text-only LLM then receives $C_i$, the question $q_i$, and four answer options, and outputs a brief rationale, an observation grounded in $C_i$, and the selected option.
When the caption provides no relevant evidence, the LLM may output the special token \texttt{unknown}.
We report overall accuracy, per-dimension accuracy, bootstrap 95\% confidence intervals, and the no-evidence rate, defined as the fraction of \texttt{unknown} outputs.
\section{Experiments}
\label{sec:experiments}

We evaluate the efficacy of the CodecCap framework by indirectly assessing the quality of its synthesized supervision. Specifically, we fine-tune a captioning model on the CodecVDC-100K dataset and evaluate it on the VidCapQA benchmark, operating under the hypothesis that higher question-answering accuracy reflects a superior preservation of fine-grained visual evidence.

\paragraph{CodecVDC-100K: Datasets.} Applying the proposed pipeline to a diverse YouTube video pool yields CodecVDC-100K, a large-scale dense video captioning dataset comprising $118{,}257$ videos ($7{,}962$ hours). Each video features a comprehensive four-level hierarchical annotation: a single video-level narrative, a median of $6$ scene-aligned segments, an anchor caption of approximately $270$ words per segment, and roughly $225$ one-second residual captions per video ($\sim$$25$ words each) explicitly documenting dynamic visual changes. As shown in Table~\ref{tab:vdc_dataset_compare}, CodecVDC-100K is the first public dataset to provide this complete anchor--residual--scene--video hierarchy, which is essential for training structured dense captioning models. We extract its residual subset as the supervised fine-tuning (SFT) data for subsequent evaluations.

\begin{table}[t]
\centering
\scriptsize
\setlength{\tabcolsep}{2.5pt}
\renewcommand{\arraystretch}{1.1}
\begin{tabularx}{\linewidth}{@{}>{\raggedright\arraybackslash}Xccccccc@{}}
\toprule
\multirow{2}{*}{\textbf{Dataset}} & \multirow{2}{*}{\textbf{\#Vid.}} & \multirow{2}{*}{\textbf{Hrs}} & \multicolumn{4}{c}{\textbf{Caption Levels}} & \multirow{2}{*}{\textbf{Pub.}} \\
\cmidrule(lr){4-7}
 & & & Anc & Res & Sc & Vid & \\
\midrule
ShareGPT4Video~\citep{chen2024sharegpt4video}    & $40$K   & $291$       & \checkmark & $\sim$    & \ding{55}  & \checkmark & \checkmark \\
Vript-CAP~\citep{yang2024vript}         & $12$K   & $1{,}300$   & \ding{55}  & \ding{55} & \checkmark & \ding{55}  & \checkmark \\
MiraData~\citep{ju2024miradata}          & $330$K  & $16$K       & \ding{55}  & \ding{55} & \checkmark & \ding{55}  & \checkmark \\
Tarsier2-Recap~\citep{zhang2025tarsier2}    & $585$K  & $1{,}972$   & \ding{55}  & \ding{55} & \ding{55}  & \checkmark & \checkmark \\
LVD-2M~\citep{ju2024lvd2m}            & $2$M    & $11$K       & \ding{55}  & \ding{55} & \checkmark & \checkmark & \checkmark \\
ViMix-14M~\citep{yang2025vimix}         & $13.7$M & $22.8$K     & \ding{55}  & \ding{55} & \ding{55}  & \checkmark & \checkmark \\
\midrule
\rowcolor[RGB]{250,239,239}
\textbf{CodecVDC-100K (ours)} & $\mathbf{118}$\textbf{K} & $\mathbf{7{,}962}$ & \checkmark & \checkmark & \checkmark & \checkmark & \checkmark \\
\bottomrule
\end{tabularx}
\caption{\label{tab:vdc_dataset_compare}Comparison of dense video captioning datasets. CodecVDC-100K uniquely provides a complete four-level hierarchy---anchor (Anc), residual (Res), scene (Sc), and video (Vid)---to support structured multi-granularity supervision ($\sim$ indicates partial coverage).}
\end{table}

\paragraph{VDCTalker: Model.} We introduce VDCTalker, developed by fine-tuning the Qwen3.5-35B-A3B base model~\citep{qwen3.5} ($35$B total, $3$B activated parameters) exclusively on the CodecVDC-100K. The model is trained using ms-swift~\cite{zhao2024swiftascalablelightweightinfrastructure} on 64 Nvidia H800 GPUs with a learning rate of 2e-5 and a global batch size of 1024. Evaluation strictly follows the VidCapQA holistic direct-captioning protocol, generating a single comprehensive caption per video to be queried by the evaluation module (see \S\ref{sec:vidcapqa_eval}).

\paragraph{Baselines.} We benchmark VDCTalker against four configurations: (i) the base Qwen3.5-35B-A3B model (to isolate the SFT impact); (ii) Gemini-3.1 Pro-Preview (a state-of-the-art proprietary model; overall accuracy only); (iii) Seed 2.0 Pro~\citep{seed2.0} (a larger-capacity holistic baseline); and (iv) Seed 2.0 Pro with a per-second prompt. This final variant evaluates one-second windows independently, establishing an empirical upper bound for temporal redundancy despite severe computational inefficiency.

\section{Results and Discussion}
\label{sec:results}

\begin{figure*}[!tp]
\centering
\includegraphics[width=\linewidth]{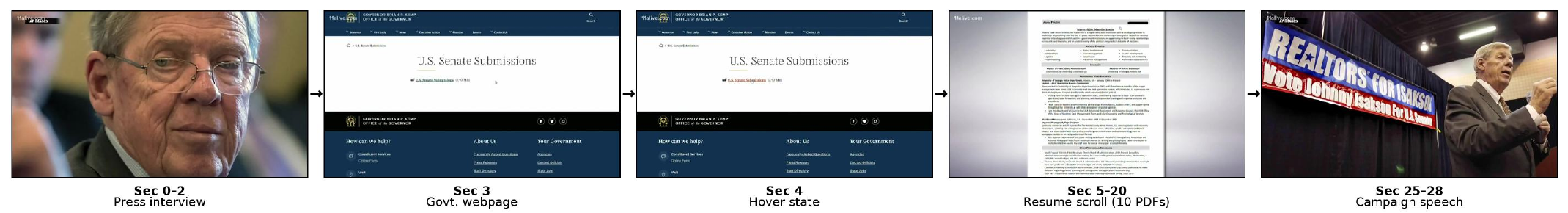}

\vspace{4pt}

\setlength{\tabcolsep}{4pt}
\renewcommand{\arraystretch}{1.0}
\scriptsize
\begin{tabular}{@{}p{0.475\linewidth} p{0.475\linewidth}@{}}
\toprule
\multicolumn{1}{c}{\textbf{Per-second baseline}} & \multicolumn{1}{c}{\textbf{Pipeline output (anchor + residuals)}} \\
\midrule
\textbf{[Sec 0]} Close-up of an elderly white man, thin-rimmed glasses, wrinkles, gazing right. Soft lighting, blurred background. ``11alive.com / AP IMAGES'' watermark.
&
\textbf{[Anchor]} An elderly man holding a silver mesh microphone, charcoal-gray suit, white shirt, brown polka-dot tie. Banner ``REALTORS\textsuperscript{\textregistered} FOR ISAKSON / Vote Johnny Isakson For U.S.\ Senate''.
\\
\textbf{[Sec 1--2]} Same close-up re-described each second: glasses, wrinkles, gaze, watermark.
&
\textbf{[Sec 1--2]} \textcolor{gray}{No visible change.}
\\
\textbf{[Sec 3]} Cuts to a webpage. ``GOVERNOR BRIAN P.\ KEMP''; heading ``U.S.\ Senate Submissions''; paperclip icon; cursor on right.
&
\textbf{[Sec 3]} Cuts to the Governor Kemp page with ``U.S.\ Senate Submissions'' and a downloadable file link.
\\
\textbf{[Sec 4]} Same webpage; cursor moves to the link; underline appears.
&
\textbf{[Sec 4]} Hover on the link turns the text red and adds an underline.
\\
\textbf{[Sec 5--20]} Document scrolls through resumes of Adam Fouche, Andrew Heilman, \dots, Melody T. McCloud---each described independently as a new page with name, sections, and bullets.
&
\textbf{[Sec 5--20]} Adam Fouche $\to$ Andrew Heilman $\to$ \dots $\to$ Melody T. McCloud: a temporally ordered stream of replacement events.
\\
\textbf{[Sec 21--24]} Returns to the elderly close-up; same description repeats for four seconds.
&
\textbf{[Sec 21--24]} \textcolor{gray}{No visible change.}
\\
\textbf{[Sec 25--30]} Cuts to a stage shot, then a wide studio shot---each scene re-described from scratch.
&
\textbf{[Sec 25--30]} Cuts to the stage shot, then to the TV studio with two anchors in front of a video wall.
\\
\bottomrule
\end{tabular}

\caption{\label{fig:case_study}Case study on a $31$-second multi-scene news clip from VidCapQA. \textbf{Top:} representative frames along the timeline. \textbf{Bottom:} side-by-side comparison of a per-second baseline and the CodecCap pipeline output (anchor + residuals).}
\end{figure*}

\subsection{Pipeline Quality}
As detailed in Table~\ref{tab:vidcapqa_results}, fine-tuning on the CodecVDC-100K residual subset yields a $5.1\%$ absolute gain in overall accuracy, improving $12$ of $14$ dimensions. Crucially, the largest gains concentrate on dynamics-centric capabilities (e.g., \textit{trajectory}, \textit{action recognition}), indicating that the anchor--residual factorization---rather than mere data scaling---drives the enhanced temporal reasoning. Despite activating only $3$B parameters, VDCTalker outperforms Gemini-3.1 Pro-Preview and rivals Seed 2.0 Pro, surpassing the latter in six dynamic dimensions. VDCTalker's lag in static attribute recognition and holistic understanding stems directly from its constrained parameter budget. While the per-second Seed variant achieves higher accuracy, it incurs severe token inflation by repeatedly redescribing static content. Finally, minor regressions in \textit{speed} and \textit{state change} highlight that residual-only supervision inherently under-emphasizes persistence-oriented attributes---a trade-off easily mitigated by co-training with our pipeline's scene- and video-level captions.

\begin{table}[!t]
\centering
\scriptsize
\setlength{\tabcolsep}{2.5pt}
\renewcommand{\arraystretch}{1.05}
\begin{tabularx}{\linewidth}{@{}>{\raggedright\arraybackslash}X c >{\columncolor[RGB]{250,239,239}}c >{\columncolor[RGB]{250,239,239}}r c c@{}}
\toprule
\textbf{Capability} & \textbf{Base} & \textbf{Ours} & \textbf{$\Delta$} & \textbf{Seed} & \textbf{Seed$+$sec} \\
\midrule
\textbf{Overall}         & $44.4$ & $49.5$ & $\mathbf{+5.1}$ & $50.9$ & $57.6$ \\
\midrule
\multicolumn{6}{@{}l}{\textit{\small Dynamics-centric capabilities}} \\
Trajectory               & $15.5$ & $\mathbf{31.0}$ & $\mathbf{+15.5}$ & $25.7$ & $35.7$ \\
Action recognition       & $50.0$ & $\mathbf{62.0}$ & $\mathbf{+12.0}$ & $54.9$ & $61.1$ \\
Temporal sequence        & $33.8$ & $45.8$ & $\mathbf{+12.0}$ & $51.4$ & $58.3$ \\
Direction                & $42.2$ & $51.4$ & $\mathbf{+9.2}$  & $66.2$ & $66.7$ \\
Rotation                 & $16.9$ & $\mathbf{23.2}$ & $\mathbf{+6.3}$  & $14.5$ & $26.9$ \\
Reasoning                & $21.1$ & $26.8$ & $\mathbf{+5.7}$  & $29.0$ & $47.8$ \\
Object tracking          & $64.2$ & $\mathbf{69.0}$ & $\mathbf{+4.8}$  & $65.1$ & $74.3$ \\
Camera movement          & $42.2$ & $\mathbf{45.1}$ & $\mathbf{+2.9}$  & $42.2$ & $48.6$ \\
Temporal grounding       & $62.0$ & $\mathbf{62.5}$ & $\mathbf{+0.5}$  & $58.7$ & $63.9$ \\
\midrule
\multicolumn{6}{@{}l}{\textit{\small Static / persistence-oriented capabilities}} \\
Attribute recognition    & $59.1$ & $62.0$ & $\mathbf{+2.9}$  & $73.1$ & $73.2$ \\
Holistic understanding   & $71.8$ & $73.2$ & $\mathbf{+1.4}$  & $78.6$ & $69.0$ \\
Counting                 & $36.9$ & $38.1$ & $\mathbf{+1.2}$  & $41.0$ & $48.6$ \\
State change             & $60.6$ & $59.1$ & $-1.5$           & $69.0$ & $76.1$ \\
Speed                    & $45.8$ & $41.7$ & $-4.1$           & $44.4$ & $53.5$ \\
\bottomrule
\end{tabularx}
\caption{\label{tab:vidcapqa_results}Performance evaluation on the VidCapQA benchmark. Capabilities are categorized into \textit{dynamics-centric} and \textit{static/persistence-oriented} groups, sorted by the absolute gain ($\Delta$) over the base Qwen3.5-35B-A3B model. \textbf{Bold} entries highlight dimensions where VDCTalker outperforms Seed 2.0 Pro. Gemini-3.1 Pro-Preview yields an overall accuracy of $47.9\%$ (per-dimension data omitted due to unavailability).}
\end{table}

\subsection{Qualitative Analysis}
\label{sec:exp_case}

Figure~\ref{fig:case_study} contrasts the CodecCap output with a per-second baseline on a $31$-second multi-scene news clip, illustrating three key architectural advantages: \textbf{(1) Redundancy Suppression:} Anchoring static scenes and emitting ``no visible change'' for unchanged frames effectively eliminates multiple verbatim redescriptions. \textbf{(2) Explicit Event Structuring:} Camera cuts are explicitly encoded as replacement events, thereby preserving temporal logic. \textbf{(3) Micro-Event Preservation:} Sub-second state transitions are isolated within residuals, whereas they are often obscured by the verbose repetitions of per-frame baselines. Consequently, CodecCap produces a highly compact token sequence while retaining denser event signals. This qualitative efficiency mirrors the quantitative findings in Table~\ref{tab:vidcapqa_results}: anchors effectively absorb static background contexts, enabling residuals to capture the high-frequency dynamic signals critical for motion-centric reasoning. This asymmetric allocation of token budgets renders the representation highly token-efficient without compromising event-level coverage.
\section{Conclusion}
\label{sec:conclusion}

We presented CodecCap, a codec-inspired framework for dense video captioning that represents videos with semantic keyframe captions and residual temporal captions. 
We further introduced VidCapQA to evaluate the recoverability of visual evidence from captions, and constructed CodecVDC-100K, a large-scale dense captioning dataset with four-level supervision.
Experiments show that CodecCap-generated supervision improves caption recoverability over direct-captioning baselines with the same underlying VLMs. 
These results suggest that codec-style semantic residual modeling is an effective and scalable direction for building faithful video-language supervision.
\section*{Limitations}
\label{sec:limitations}

The current implementation of CodecCap is subject to two primary limitations. First, it relies exclusively on the visual modality, omitting auditory signals (e.g., speech and environmental sounds), which may yield incomplete semantic representations for audio-centric content. Nevertheless, the proposed anchor--residual factorization is intrinsically extensible, offering a pathway to model discrete audio events as temporal residuals against a stable acoustic background in future work. Second, because the residual subset of CodecVDC-100K is explicitly designed to prioritize temporal dynamics over persistent states to isolate motion-centric supervision, we observe minor performance degradation in the \textit{speed} and \textit{state change} dimensions (Table~\ref{tab:vidcapqa_results}). However, this stems from an intentional data sampling strategy rather than a fundamental flaw in the representational framework. Since persistent visual attributes are comprehensively captured by the scene- and video-level captions synthesized within the same pipeline, future iterations can seamlessly mitigate these gaps through a joint training paradigm across the complete four-level hierarchy, necessitating no additional data collection.

\bibliography{references}

\appendix

\end{document}